\documentclass[3p,times,procedia]{elsarticle}
\usepackage{ecrc}
\usepackage[bookmarks=false]{hyperref}
    \hypersetup{colorlinks,
      linkcolor=blue,
      citecolor=blue,
      urlcolor=blue}

\volume{00}

\firstpage{1}

\journalname{Procedia Computer Science}

\runauth{Author name}

\jid{procs}

\usepackage{subfigure}
\usepackage{amsmath,amssymb,amsfonts}
\usepackage{algorithmic}
\usepackage{graphicx}

\usepackage[figuresright]{rotating}

\begin{document}
\begin{frontmatter}



\dochead{International Conference on Machine Learning and Data Engineering}

\title{Enhancing Image Authenticity Detection: Swin Transformers and Color Frame Analysis for CGI vs. Real Images}


\author[a]{Preeti Mehta} 
\author[b]{Aman Sagar}
\author[b]{Suchi Kumari\corref{cor1}}

\address[a]{Department of Computer Science and Engineering, National Institute of Technology Delhi, Delhi, India}
\address[b]{Department of Computer Science and Engineering, Shiv Nadar Institute of Eminence, Delhi-NCR, India}

\begin{abstract}
The rapid advancements in computer graphics have greatly enhanced the quality of computer-generated images (CGI), making them increasingly indistinguishable from authentic images captured by digital cameras (ADI). This indistinguishability poses significant challenges, especially in an era of widespread misinformation and digitally fabricated content. This research proposes a novel approach to classify CGI and ADI using Swin Transformers and preprocessing techniques involving RGB and CbCrY color frame analysis. By harnessing the capabilities of Swin Transformers, our method foregoes handcrafted features instead of relying on raw pixel data for model training. This approach achieves state-of-the-art accuracy while offering substantial improvements in processing speed and robustness against joint image manipulations such as noise addition, blurring, and JPEG compression. Our findings highlight the potential of Swin Transformers combined with advanced color frame analysis for effective and efficient image authenticity detection.
\end{abstract}

\cortext[cor1]{Corresponding author. Suchi Kumari}
\end{frontmatter}

\email{suchi.singh24@gmail.com}

\vspace*{-6pt}

\section{Introduction}
\label{introduction}
 
\begin{figure}
	\centering
	\includegraphics[width =8cm,height =1.5cm]{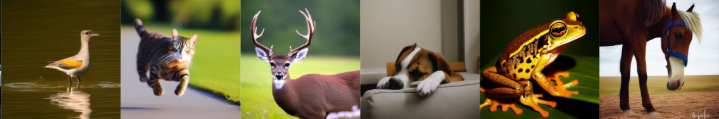}
	
\caption{Example of few Computer Generated (CG) images from dataset CIFAKE-10. The images shows the difficulty in distingushing between two classes of images with naked eye. }
\label{fig1}
	
\end{figure}

Natural images (NIs) represent real-world scenes, while computer graphics advances have created highly realistic computer-generated (CG) images. As these CG images become more visually plausible, distinguishing them from NIs has become a critical challenge in digital forensics. The main difficulty lies in the goal of computer graphics: to achieve photorealism that matches the authenticity of natural images. This challenge is exemplified by images that, at first glance, may be indistinguishable from photographs taken with digital cameras. Figure \ref{fig1} illustrate some CG images from CIFAKE-10 image dataset \cite{bird2024cifake}.

Traditional approaches to distinguishing between authentic and computer-generated images can be categorized into subjective and objective methods. Subjective methods rely on psychophysical experiments, while objective methods depend on the statistical and intrinsic properties of images. Conventional objective methods often involve creating sophisticated, discriminative, handcrafted features and then training classifiers such as support vector machines (SVM) or ensemble classifiers\cite{lyu2005realistic,chen2007identifying}. While these methods can perform well on simpler datasets, they are unable to carch up with complex datasets containing images from diverse sources. Recent advancements in neural networks and vision transformers have transformed computer vision and pattern recognition, offering a powerful alternative to handcrafted feature-based methods. Convolutional neural networks (CNNs) are particularly effective for complex datasets because they automatically learn multiple levels of representation from data in an end-to-end manner. Inspired by this success, researchers have begun applying CNNs to multimedia security and digital forensics \cite{ni2019evaluation,quan2018distinguishing,yao2022cgnet}.

This research proposes a novel approach to distinguishing between computer-generated and real digital camera images using Swin Transformers and color frame analysis. Swin Transformers, known for their ability to handle various image recognition tasks effectively, eliminate the need for handcrafted feature extraction by directly utilizing raw pixel data. Our method incorporates preprocessing techniques involving RGB and CbCrY color frames to enhance classification. Additionally, we employ TSNE plots to visualize the feature spaces and demonstrate the effectiveness of our approach. 
Our contributions include:
\begin{itemize}
	\item We are introducing a Swin Transformer-based framework for classifying CGI and ADI.
	\item We use color frame analysis in RGB and CbCrY spaces to improve preprocessing.
	\item We have achieved state-of-the-art accuracy of 98\% for the RGB format images.
\end{itemize}

The layout of the paper is as follows: Section \ref{introduction} discusses the importance of distinguishing computer-generated images from real ones in image forensics. Section \ref{relatedwork} reviews traditional methods and recent deep learning advancements. Section \ref{method} describes our approach using the Swin Transformer for image classification, including preprocessing and model architecture. Section \ref{result} details the experimental results, dataset usage, setups, and performance evaluations. Section \ref{conclusion} summarizes the findings, acknowledges limitations, and suggests future research directions.


\section{Related Work}
\label{relatedwork}

Various methodologies for detecting computer-generated (CG) images focus on feature extraction and classification. One approach involves identifying statistical artefacts left by graphical computer generated modules and using threshold-based evaluation. Ng \textit{et al.} \cite{ng2005physics} achieved 83.5\% accuracy by designing object geometry features to distinguish physical disparities between photographic and CG images. Wu \textit{et al.} \cite{wu2006detecting} used Gabor filter as feature discriminator for visual features like color, edge, saturation, and texture. Dehnie \textit{et al.} \cite{dehnie2006digital} emphasized differences in image acquisition between digital cameras and generative algorithms, creating features based on residual images from wavelet-based denoising filters. Texture-based methods have also been employed for CG and photographic (PG) classification. Li \textit{et al.} \cite{li2014distinguishing} achieved 95.1\% accuracy using uniform gray-scale invariant LBP feature vectors. Despite their interpretability, hand-crafted feature-based methods are limited by manual design constraints and feature description capacities.

In response to the shortcomings of hand-crafted features and machine learning, recent research has leaned towards leveraging deep learning methods for improved detection performance. For instance,
Mo \textit{et al.} \cite{mo2018fake} presented a CNN-based method that concentrates on high-frequency components, achieving an average accuracy exceeding 98\%. Meena \textit{et al.} \cite{meena2021distinguishing} presented a two-stream convolutional neural network that integrates a pre-trained VGG-19 network for trace learning and uses high-pass filters to highlight noise-based features. Yao \textit{et al.} \cite{yao2022cgnet} introduced a novel approach utilizing VGG-16 architecture combined with a Convolutional Block Attention Module, achieving an impressive accuracy of 96\% on the DSTok dataset. Furthermore, Chen \textit{et al.} \cite{chen2021locally} presented an enhanced Xception model tailored for locally generated face detection in GANs. Liu \textit{et al.} \cite{liu2022detecting} introduced a method focusing on authentic image noise patterns for detection. These recent advancements underscore the growing trend of applying sophisticated deep-learning architectures to enhance the detection accuracy of computer-generated images, addressing the challenges posed by increasingly realistic synthetic imagery.

\section{Proposed Methodology}
\label{method}
\label{methodology}
\begin{figure*}[!htb]
	\centering
	
	\includegraphics[width=0.95\linewidth, height= 3.5cm]{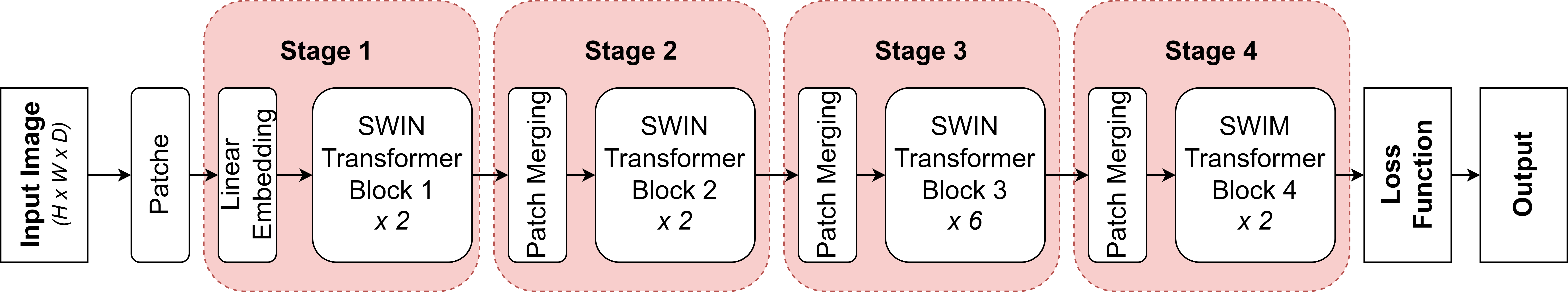}
	\caption{Architecture of the proposed Swin transformer}
	\label{fig2}
\end{figure*}

\subsection{Swin Transformer Architecture}
The Swin Transformer is a cutting-edge deep learning architecture renowned for its hierarchical representations, which efficiently capture long-range spatial dependencies in large-scale visual data \cite{mehta2024domain}. The architecture figure \ref{fig2} illustrates the hierarchical structure of the Swin Transformer, showcasing its ability to capture global context and local details simultaneously. By leveraging self-attention mechanisms and multi-layered processing, the Swin Transformer excels at learning complex patterns in visual data, making it ideal for image classification tasks. It is characterized by self-attention mechanisms. Mathematically, the self-attention mechanism of the Swin Transformer can be represented as follows:

\[
\text{Att}(\mathbb{Q, K, V}) = \text{SoftMax} \left( \frac{\mathbb{QK}^T}{\sqrt{k_v}} \right) \mathbb{V}
\]

where \( \mathbb{Q} \), \( \mathbb{K} \), and \( \mathbb{V} \) represents the matrices query, key, and value, respectively, and \( k_v \) denotes the key vectors dimensionality.

Our study employed the Swin Transformer to distinguish between computer-generated images (CGI) and authentic images. We conducted color frame analysis to enhance the model's understanding of the distinct characteristics of CGI and authentic images. By analyzing the RGB color channels and extracting meaningful features, we aimed to provide the model with valuable insights into the differences in color distribution and spatial arrangements between CGI and authentic images.

\subsection{Color Frame Analysis}

In our methodology, we leverage color frame analysis to extract discriminative features from RGB and CbCrY color frames. The process involves transforming RGB images into the CbCrY color space and computing the color features using mathematical formulations:

\[
\begin{bmatrix}
	Y \\
	Cb \\
	Cr
\end{bmatrix}
=
\begin{bmatrix}
	0.300 & 0.586 & 0.113 \\
	-0.168 & -0.328 & 0.496 \\
	0.496 & -0.414 & -0.082
\end{bmatrix}
\begin{bmatrix}
	R \\
	G \\
	B
\end{bmatrix}
+
\begin{bmatrix}
	0 \\
	128 \\
	128
\end{bmatrix}
\]

These equations represent the conversion of RGB color channels to the YCbCr color channels, where \(R\), \(G\), \(B\), \( Y \), \( Cb \), and \( Cr \) denote the red, green, blue, luminance, and chrominanceof blue and red components, respectively.

\subsection{Feature Visualization with t-SNE}

To visualize the impact of color frame analysis on feature extraction, we have utilized t-Distributed Stochastic Neighbor Embedding (t-SNE) plots. Mathematically, t-SNE minimizes the Kullback-Leibler divergence between two distributions of feature vectors in high and low-dimensional spaces. The t-SNE algorithm can be expressed as:

\[
ps_{ij} = \frac{\exp(-\lVert x_i - x_j \rVert^2 / 2 \sigma_i^2)}{\sum_{k \neq i} \exp(-\lVert x_i - x_k \rVert^2 / 2 \sigma_i^2)}
\]
\[
qs_{ij} = \frac{(1 + \lVert y_i - y_j \rVert^2)^{-1}}{\sum_{k}\sum_{k \neq l} (1 + \lVert y_k - y_l \rVert^2)^{-1}}
\]
\[
KLD(P || Q) = \sum_{i\neq j} ps_{ij} \log \frac{ps_{ij}}{qs_{ij}}
\]

Here, \( ps_{ij} \) represents the pairwise similarity between points \( x_i \) and \( x_j \) in the high-dimensional space, \( qs_{ij} \) denotes the pairwise similarity between points \( y_i \) and \( y_j \) in the low-dimensional space, and \( KLD \) represents the Kullback-Leibler divergence between \(P\) and \(Q\) distributions.

\begin{figure}[!htb]
	\centering
	\subfigure[\label{fig3a}]{\includegraphics[width=0.4\linewidth,height=4.5cm]{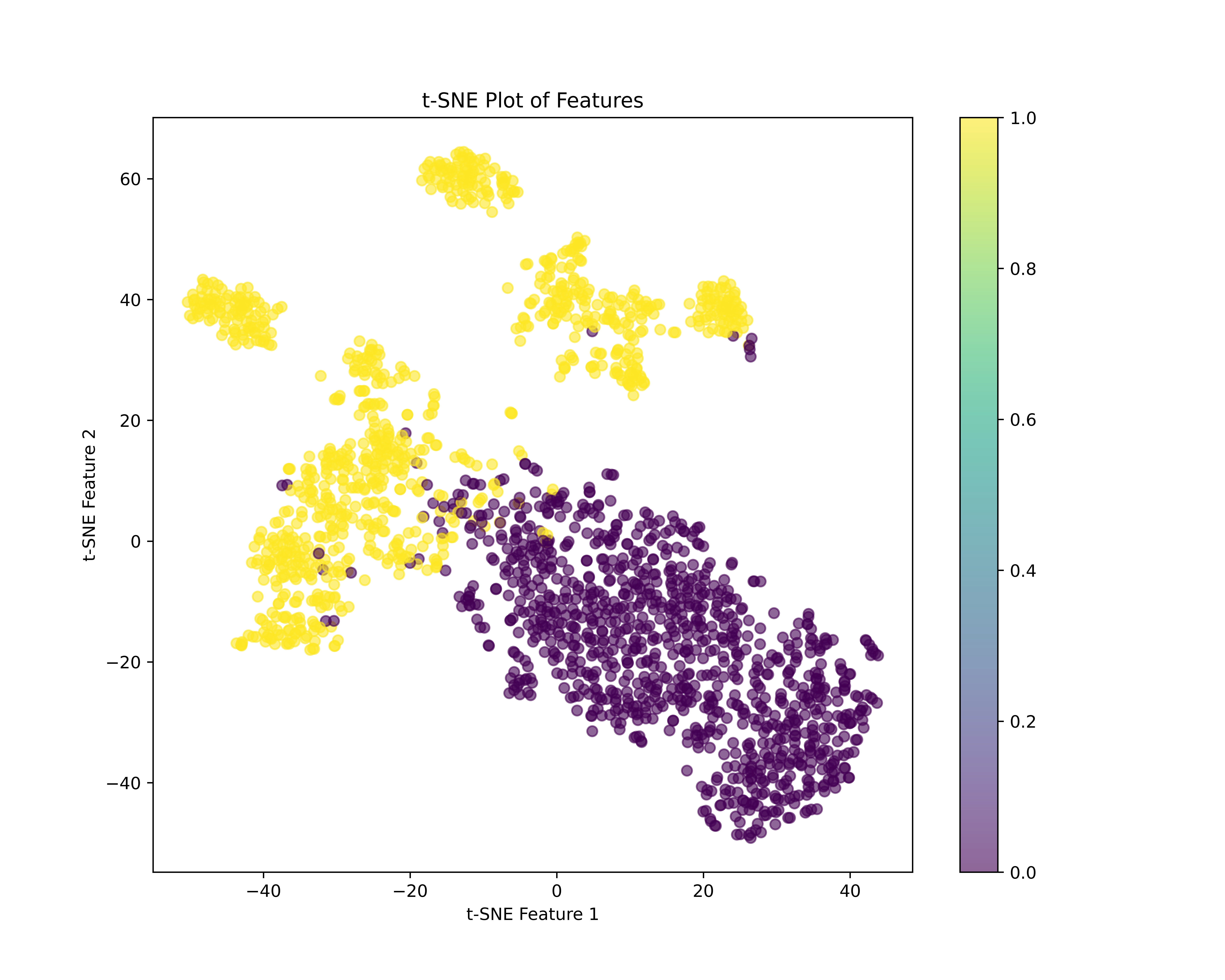}}
	\subfigure[\label{fig3b}]{\includegraphics[width=0.4\linewidth,height=4.5cm]{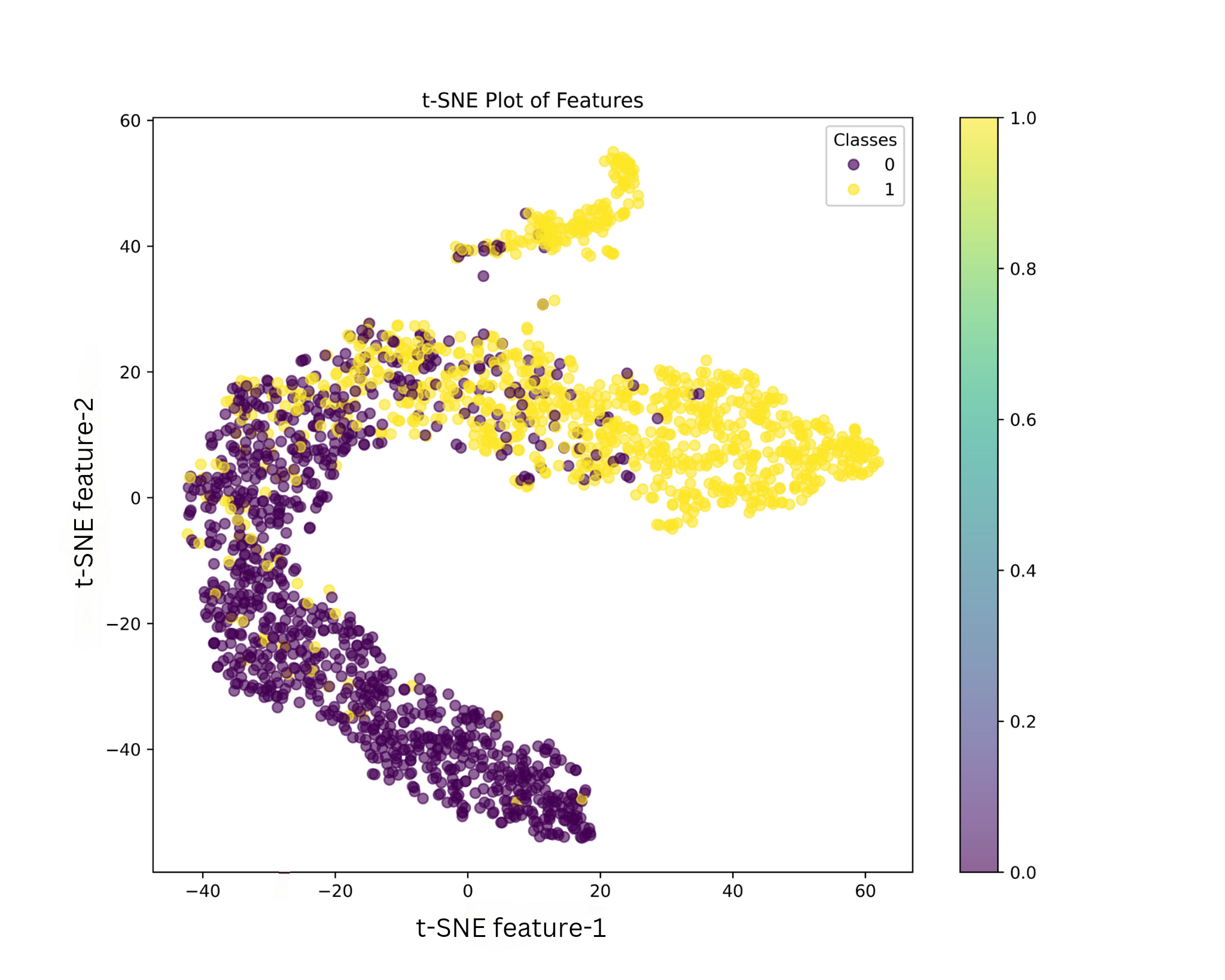}}
	\caption{The plot \ref{fig3a} illustrate the t-SNE plot of the extracted features from the Swin Tranformer for the RGB input images. The plot \ref{fig3b} illustrate the t-SNE plot of the extracted features from the Swin Tranformer for the CbCrY input images. }
	\label{fig3}
\end{figure}

Figure \ref{fig3} showcases t-SNE plots of the extracted features from the Swin transformer, demonstrating the impact of color frame analysis on feature separability. The yellow dots represents CGI extracted features and purple dots represents the ADI extracted features in two-dimension space.

\section{Results and Discussion}
\label{result}
We train the Swin Transformer using the extracted features from both RGB and CbCrY color frames, separately. The training process involves minimizing the cross-entropy loss function:

\[
\mathcal{L} = - \frac{1}{\mathbb{N}} \sum_{i=1}^{\mathbb{N}} \sum_{j=1}^{\mathbb{C}} t_{ij} \log(p_{ij})
\]

where \( \mathbb{N} \) represents the number of samples, \( \mathbb{C} \) denotes the number of classes, i.e. two for our problem, \( t_{ij} \) are the true labels, and \( p_{ij} \) are the predicted label class.
We evaluate the performance of our proposed methodology using standard classification metrics such as accuracy, precision, recall, and F1-score. These metrics measure the model's ability to correctly classify images as either CGI or authentic based on the extracted features. Our model is executed in PyTorch and runs on a Dell Inspiron 5502 with an i5 CPU processor and 16GB of RAM. The hyperparameter details are provided in Table \ref{table1}.
\begin{table}[!htbp]
	\centering
	\caption{Hperparameters}
	\label{table1}
	\begin{tabular}{|l|c|}
		\hline
		\textbf{Hyperparameters} & \textbf{Values}\\
		\hline
		Learning Rate & 0.0001\\ \hline
		Optimizer & Adam\\ \hline
		Loss function &Cross Entropy\\ \hline
		Batch Size& 32\\ \hline
		Input Image Size &224$\times$224$\times$3\\ \hline
		Normalized Mean&[0.485, 0.456, 0.406]\\ \hline
		Normalized Std &[0.229, 0.224, 0.225]\\ \hline
		Epochs & 5\\
		\hline
	\end{tabular}
\end{table}

Our study primarily utilized the CIFAKE-10 dataset \cite{bird2024cifake}, an extensive collection of computer-generated images (CGI) and authentic digital images (ADI). This dataset contains over one million images, divided into two categories: \lq\lq Fake\rq\rq and \lq\lq Real\rq\rq. The images in CIFAKE-10 were meticulously curated to ensure high-quality AI-generated images with reliable labeling, making it an excellent choice for our proposed methodology. The class division in the dataset is detailed in Table \ref{table2}. 

\begin{table}[!htbp]
	\centering
	\caption{Dataset Information}
	\label{table2}
	\begin{tabular}{|l|c|c|c|c|}
	\hline
	\textbf{Class} & \textbf{Generator} & \textbf{Training} & \textbf{Validation} & \textbf{Testing}\\ \hline
	\textbf{Fake}&Diffusion Model&45,000&5,000&10,000\\ \hline
	\textbf{Real}&N/A&45,000&5,000&10,000\\ \hline
	\textbf{Total}&N/A&90,000&10,000&20,000\\ \hline
	\end{tabular}
\end{table}

\begin{figure}[!htb]
    \centering
	\subfigure[\label{fig4a}]{\includegraphics[width=0.4\linewidth,height=4.5cm]{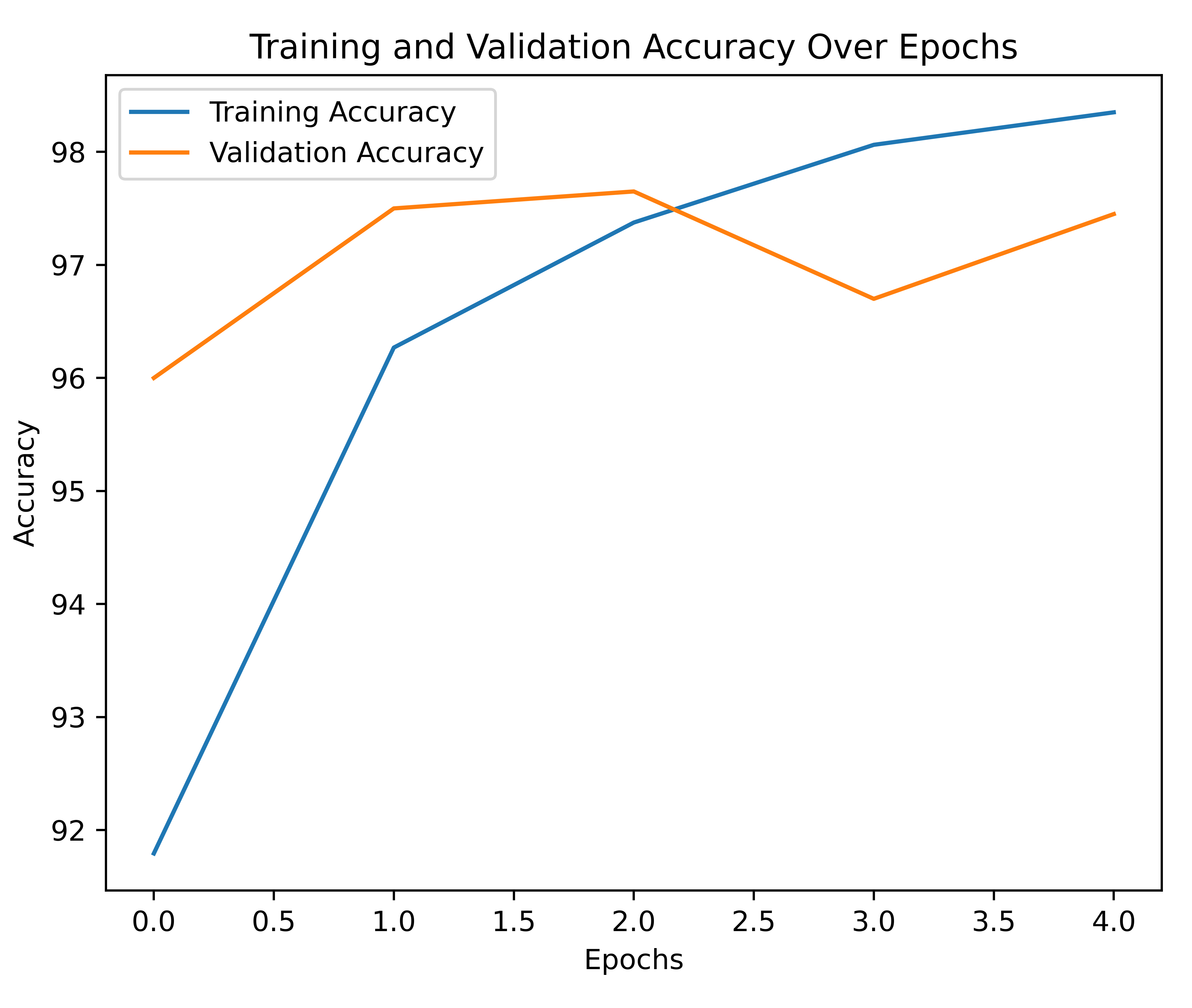}}
	\subfigure[\label{fig4b}]{\includegraphics[width=0.4\linewidth,height=4.5cm]{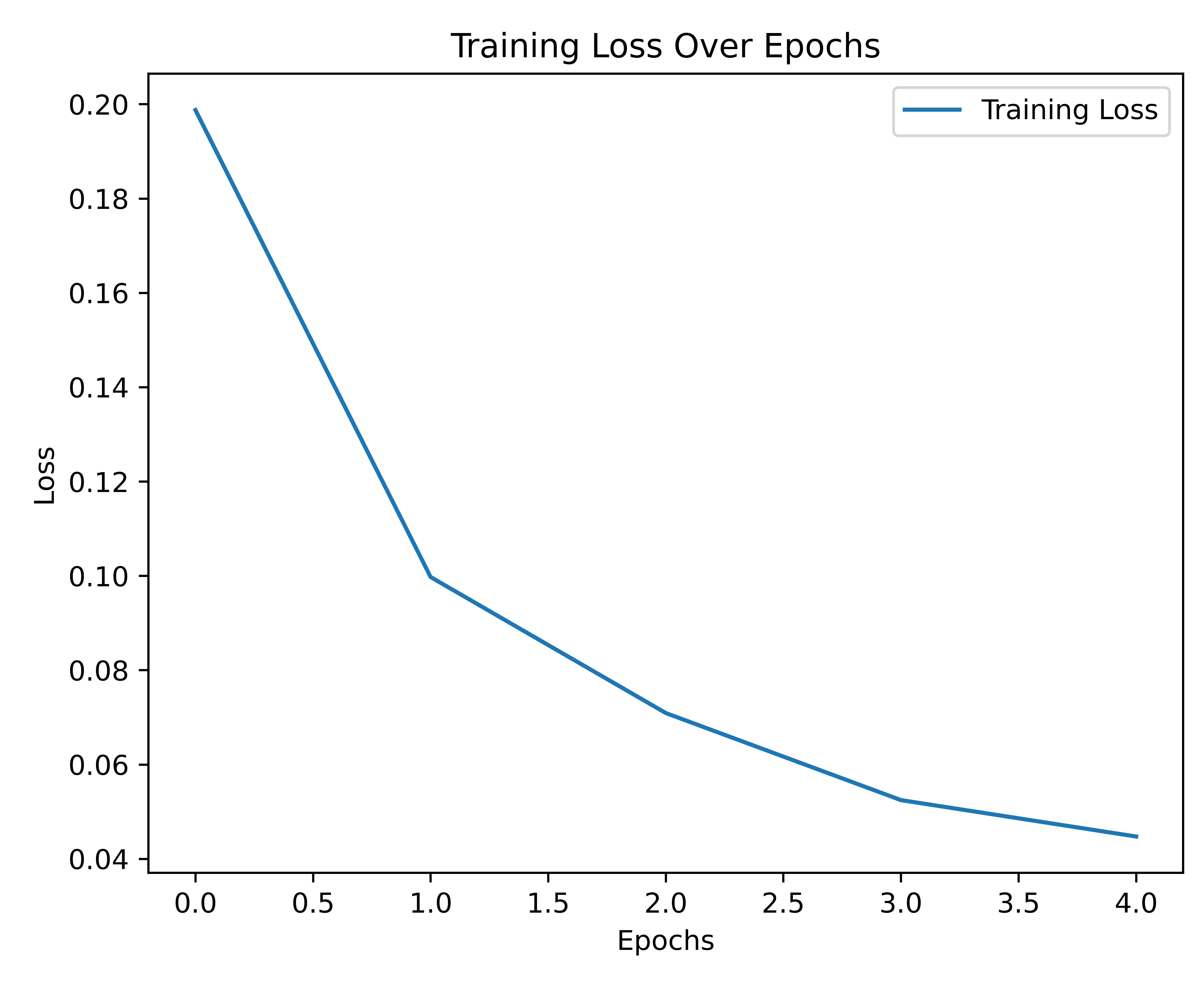}}\\
	\subfigure[\label{fig4c}]{\includegraphics[width=0.4\linewidth,height=4.5cm]{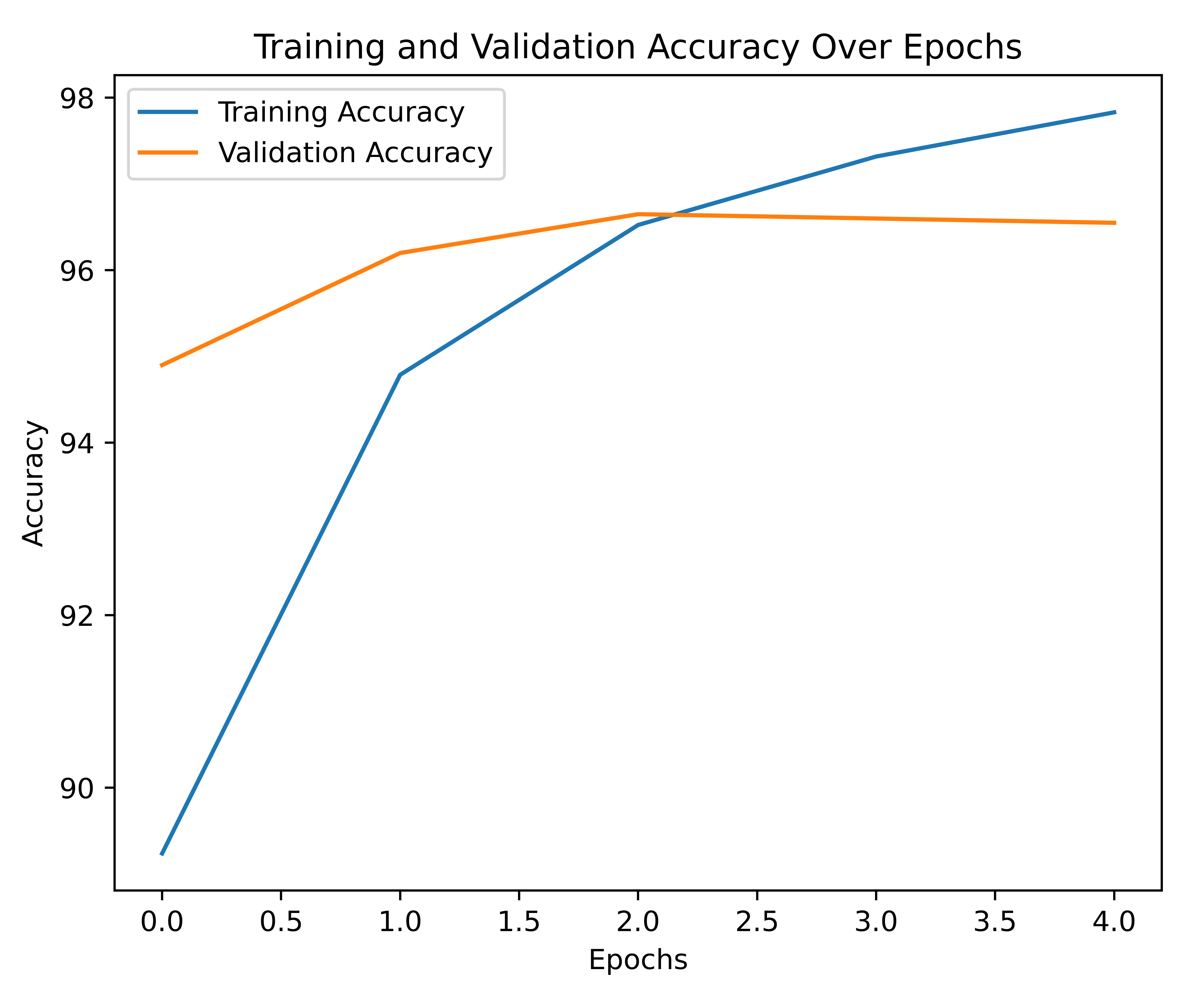}}
	\subfigure[\label{fig4d}]{\includegraphics[width=0.4\linewidth,height=4.5cm]{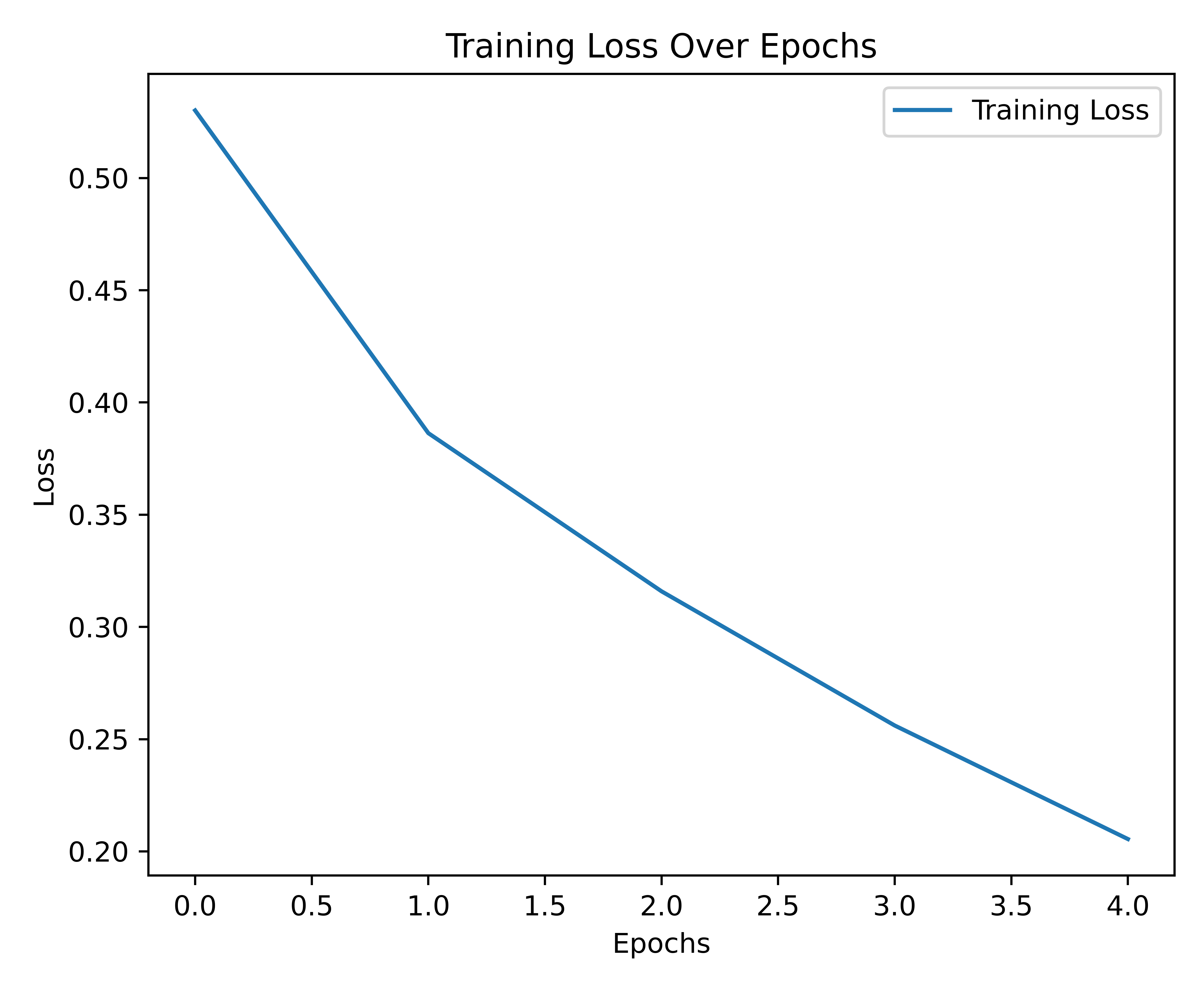}}
	\caption{The plot \ref{fig4a} and \ref{fig4b} illustrate the training and validation accuracy and loss values, respectively from the proposed model for the RGB input images for five epochs. The plot \ref{fig4c} and \ref{fig4d} illustrate the training and validation results for the CbCrY input images for five epochs. }
	\label{fig4}
\end{figure}

\begin{figure}[!htb]
	\centering
	\subfigure[\label{fig5a}]{\includegraphics[width=0.4\linewidth,height=4.5 cm]{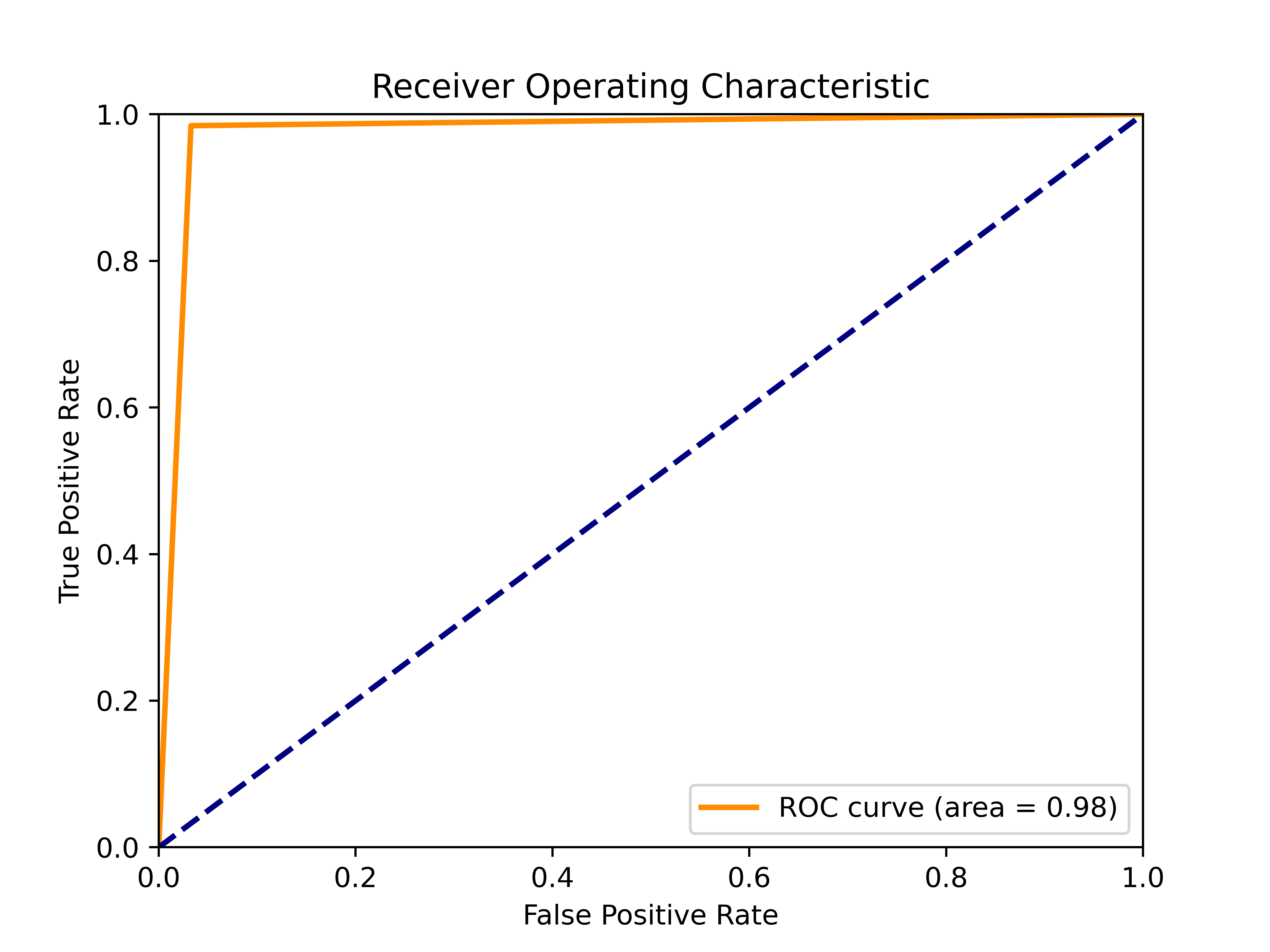}}
	\subfigure[\label{fig5b}]{\includegraphics[width=0.4\linewidth,height=4.5 cm]{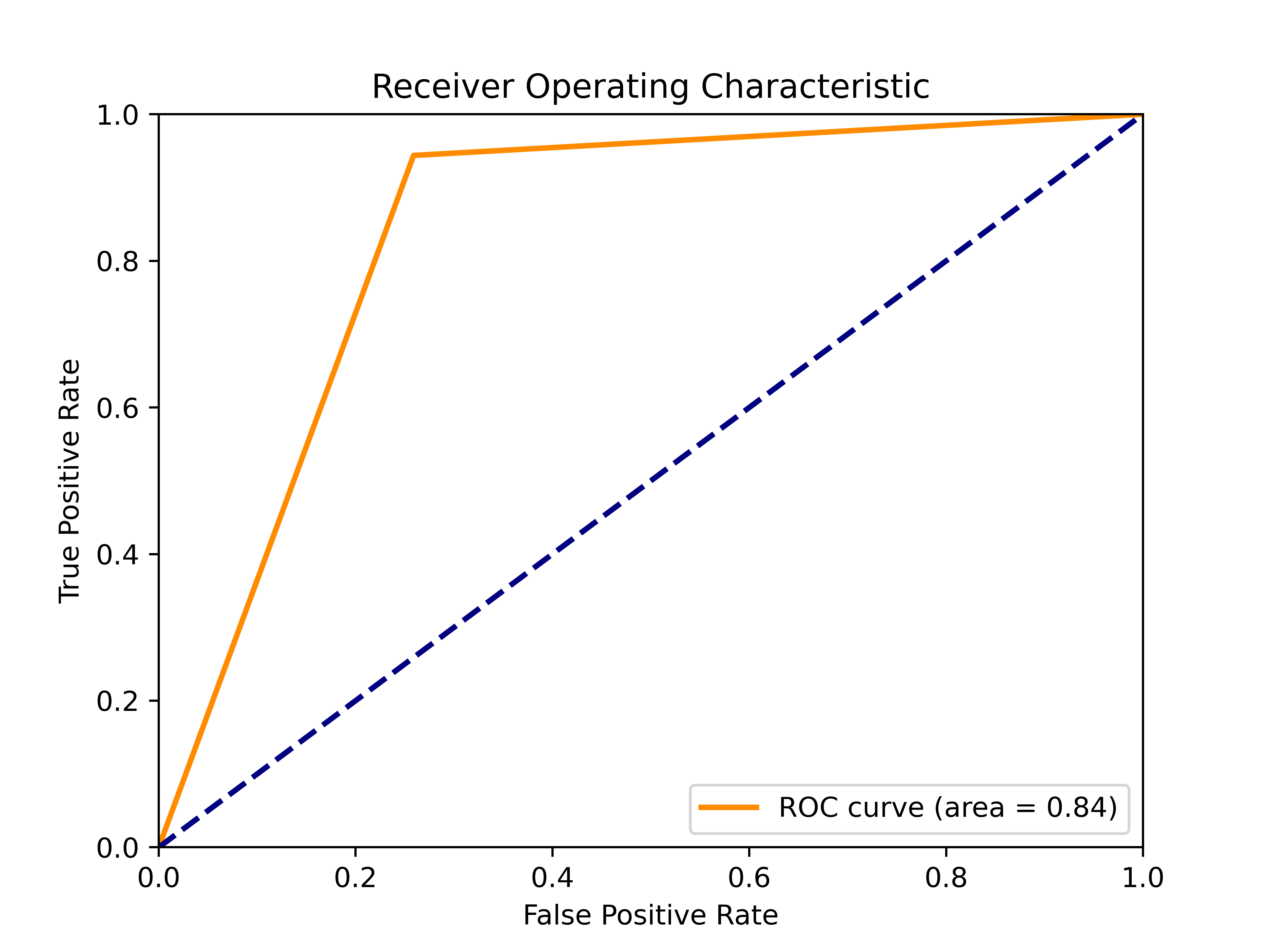}}

\caption{The ROC plots \ref{fig5a} and \ref{fig5b} for the RGB and CbCrY format input images for the proposed model using Swin Transformer.}
\label{fig5}
\end{figure}

The figures \ref{fig4} and \ref{fig5}, illustrate the training, validation and testing evaluated results for the RGB and CbCrY format input images passes through the Swin Transformer (ST) deep learning model. The results shows that, RGB format images can be better distinguished by the transformer when compare the results obtained from the CbCrY color format images. The t-SNE plots also shows the complexities of the features for different color formats.

\begin{table}
	\centering
	\caption{Comparative Performance with other models. All evaluation parameters are in \%. Best results are in Bold.}
	\label{table3}
	\begin{tabular}{|l|c|c|c|c|}
	\hline
	\textbf{Model} & \textbf{Accuracy} & \textbf{Precision} &\textbf{Recall}&\textbf{F1-score}\\ \hline
	ResNet-50 &90.20&90.09&90.34&90.21\\ \hline
	VGG-16&89.44&89.31&89.60&89.45\\ \hline
	EfficientNet V2B0 &90.75&90.57&90.98&	90.77\\ \hline
	MobileNet V3Small &81.05&81.27&80.69&80.97\\ \hline
	Proposed Model (ST) &\textbf{98.45}&\textbf{97.15}&\textbf{97.60}&\textbf{97.37}\\ \hline
	\end{tabular}
\end{table}

We have compared our model with other state-of-the-art available deep learning neural networks. The table\ref{table3}, shows that the Swin Transformer outperform all the other neural networks by a great margin. The results mentioned for ST is for the rgb images. The accuracy achieved for the RGB images is 98\% and for the CbCrY images is 84\%. Additionally, we observe that the ResNet-50 and
EfficientNet V2B0, perform satisfactorily across the different models.

\section{Conclusion and Future Directions}
\label{conclusion}
In this research, we used the Swin Transformer model with RGB and CbCrY images to differentiate between authentic digital (ADI) and computer-generated synthetic (CGI) images. While traditional neural networks like ResNet, VGG, and MobileNet provide a solid baseline, transformer models demonstrated superior classification accuracy by effectively extracting local and global features in complex feature spaces.
Future works could focus on optimizing these models for computer generated media from different plateform like GANs, diffusion models, data augmentations etc. Additionally, examining AI's ethical implications and potential misuse such as deepFake in creating synthetic media is crucial. Our study centered on one specific dataset, domain generalization is one prospective future direction that is unexplored in this area.

\bibliographystyle{elsarticle-harv}
\bibliography{PROCS_ICMLDE2024}

\end{document}